\documentclass[11pt]{article}
\usepackage{acl2020}
\usepackage{times}
\usepackage{url}
\usepackage{latexsym}
\usepackage{amsmath}
\usepackage{graphicx}
\usepackage{color}
\usepackage{xspace}
\usepackage{booktabs}
\usepackage{listings}
\usepackage{ascii}

\aclfinalcopy

\newcommand{\cut}[1]{}
\newcommand\todo{\textcolor{red}}

\newif\ifshowcomments
\showcommentsfalse

\newcommand{\cd}[1]{{\textcolor{blue}{#1}}}
\newcommand{\justine}[1]{{\textcolor{magenta}{#1}}}
\newcommand{\ly}[1]{{\textcolor{cyan}{#1}}}
\newcommand{\jpc}[1]{{\textcolor{green}{#1}}}
\newcommand{\caleb}[1]{{\textcolor{orange}{#1}}}

\ifshowcomments
\hypersetup{draft}
\else
\renewcommand\todo[1]{}
\renewcommand{\cd}[1]{}
\renewcommand{\justine}[1]{}
\renewcommand{\ly}[1]{}
\renewcommand{\jpc}[1]{}
\renewcommand{\caleb}[1]{}
\fi

\newcommand{\xhdr}[1]{{\noindent\bfseries #1.}}

\DeclareTextFontCommand{\textascii}{\asciifamily}
\newcommand{\corpus}{\textascii{Corpus}\xspace}
\newcommand{\conversation}{\textascii{Conversation}\xspace}
\newcommand{\utterance}{\textascii{Utterance}\xspace}
\newcommand{\user}{\textascii{Speaker}\xspace}
\newcommand{\transformer}{\textascii{Transformer}\xspace}
\newcommand{\fit}{\textascii{fit()}\xspace}
\newcommand{\transform}{\textascii{transform()}\xspace}

\title{ConvoKit: A Toolkit for the Analysis of Conversations}

\author{
  Jonathan P. Chang \\
  Cornell University \\
  {\tt jpc362@cornell.edu} \\ \And
  Caleb Chiam \\
  Cornell University \\
  {\tt cc982@cornell.edu} \\ \And
  Liye Fu \\
  Cornell University \\
  {\tt liye@cs.cornell.edu} \\ \AND
  Andrew Z. Wang \\
  Stanford University \\
  {\tt anwang@cs.stanford.edu} \\ \And
  Justine Zhang \\
  Cornell University \\
  {\tt jz727@cornell.edu} \\ \And 
  Cristian Danescu-Niculescu-Mizil \\
  Cornell University \\
  {\tt cristian@cs.cornell.edu}
}

\begin{document}

\maketitle
\begin{abstract}

This paper describes the design and functionality of ConvoKit, an open-source toolkit for analyzing conversations and the social interactions embedded within.
ConvoKit provides an unified framework for representing and manipulating conversational data, as well as a large and diverse collection of conversational datasets.
By providing an intuitive interface for exploring and interacting with conversational data, this toolkit lowers the technical barriers for the broad adoption of computational methods for conversational analysis.

\end{abstract}

\section{Introduction}
\label{sec:intro}

The NLP community has benefited greatly from the public availability of standard toolkits, such as NLTK \cite{bird_natural_2009},
 StanfordNLP \cite{qi_universal_2018}, 
spaCy \cite{honnibal_spacy_2020},
or scikit-learn \cite{pedregosa_scikit-learn_2011}.
These toolkits allow researchers to focus on developing new methods rather than on re-implementing existing ones, and encourage reproducibility.  Furthermore, by lowering the technical entry level, 
they 
facilitate the export of NLP techniques to other fields.

Although much of natural language is produced in the context of conversations, none of the existing public NLP toolkits are specifically targeted at the analysis of conversational data.
In this paper, we introduce ConvoKit ({\url{https://convokit.cornell.edu}}), a 
Python package that provides a
unified open-source framework for computationally analyzing  conversations and the social interactions 
taking place
 within, as well as a large collection of conversational data in a compatible format.

In designing a toolkit 
for analyzing conversations, we start from 
some basic guiding principles.
Firstly, conversations are more than mere `bags of utterances', so we must capture what connects utterances into meaningful interactions. This translates into native support of reply and tree structure as well as other dependencies across utterances.

Secondly, conversations are inherently social.
People often engage in multiple conversations, and how
 we understand interactions is contingent on what we know about the respective interlocutors.  Similarly, the way we understand each speaker is contingent on their entire conversational history.
Thus, a conversational analysis toolkit must allow for the integration of speaker information and behaviors across different conversations. 

Thirdly, conversations occur in vastly different contexts, from dyadic face-to-face interactions, to discussions and debates in institutional settings, to online group discussions, and to large-scale threaded  discussions on social media.
This means that the toolkit must offer a level of abstraction that supports different 
 interaction formats.
Finally, since conversational data is key to
many
social science fields (e.g. political science, sociology, social psychology), the framework should be accessible to a broad audience: not only experienced NLP researchers, but anyone with questions about conversations who may not necessarily have a high degree of 
NLP expertise.

In this paper, we describe how these principles guided our design of ConvoKit's 
framework
 architecture (Section \ref{sec:design}), describe some of the 
analysis methods (Section \ref{sec:transformers}) and datasets (Section \ref{sec:data}) included in ConvoKit,
and conclude with some high-level remarks on future developments (Section \ref{sec:discussion}).

\section{Framework Architecture}
\label{sec:design}

\begin{figure}
    \centering
    \includegraphics[width=0.9\linewidth]{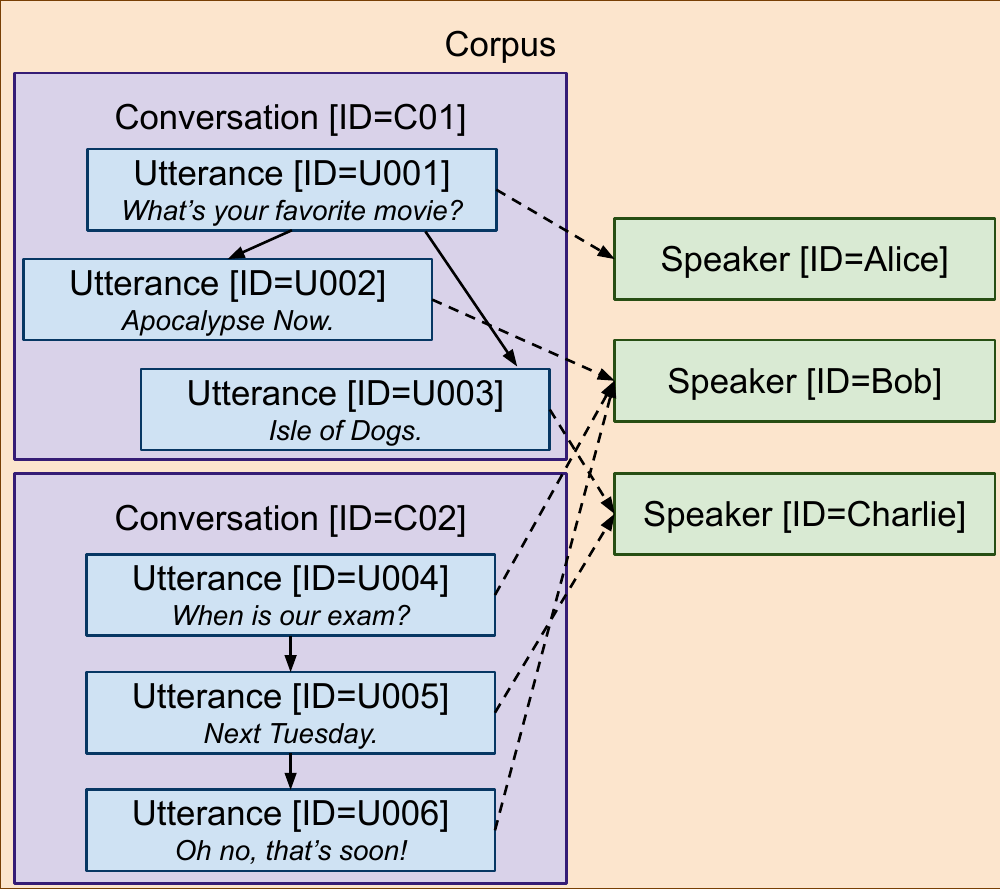}
    \caption{Visualization of the relationship between the four core classes of the \corpus hierarchy. Solid arrows denote reply-to relationships between \utterance{s}, while dashed arrows denote attribution of each \utterance to its authoring \user.}
    \label{fig:architecture}
\end{figure}

The current state of the software and data ecosystem for conversational
research is fragmented: popular conversational datasets 
are each distributed in different data formats, each using their own task-specific schemas, while similarly, code for reproducing various conversational methods 
tends to be ad-hoc with no guarantee of overlapping functionality or 
cross-compatibility.
This combination of factors poses a barrier to both reproducibility and broader adoption.
To address these issues,
a \emph{unified} framework for 
analyzing conversations
must provide
both
a standardized format for \emph{representing} any conversational data, and a general language for describing \emph{manipulations} of said data.
Furthermore, as described in Section \ref{sec:intro}, the representation 
must go beyond a mere ``bag-of-utterances'' and natively capture the structure of conversations, while the language of manipulations must be expressive enough to describe actions at different levels of the conversation: individual utterances, entire conversations, speakers in and across conversations, and arbitrary combinations of the above.

These criteria directly lead to the two core abstractions underlying ConvoKit: the \corpus, representing a collection of one or more conversations, and the \transformer, representing some action or computation that can be done to a \corpus.
To draw an analogy to language, \corpus objects are the nouns of ConvoKit, while \transformer{s} are the verbs.

\xhdr{Representing conversational data}
The main data structure for organizing conversational data in ConvoKit is the \corpus, which forms the top of a hierarchy of classes representing different levels of a conversation (Figure \ref{fig:architecture}):
A \corpus is a collection of \conversation{s}, each \conversation is made up of one or more \utterance{s}, and each \utterance is attributed to exactly one \user (but each \user can own multiple \utterance{s}).
\conversation{s}, \utterance{s} and \user{s} are identified by unique IDs.
Conversation structure is represented by the \textascii{reply\_to} field of the \utterance class, which specifies the ID of the other \utterance it replies to (i.e., its parent node in the conversation tree). ConvoKit leverages the relationships between \utterance{s}, \user{s}, and \conversation{s} to provide rich navigation of a \corpus, such as tree traversal of \utterance{s} within a \conversation or chronological iteration over all of a \user's \utterance history.

\xhdr{Custom metadata}
Objects in the \corpus hierarchy contain some basic information that is generally useful for 
most operations on conversational data, such as the text content and timestamp of each \utterance.
However, any use of ConvoKit beyond basic analyses will likely require 
additional task-specific 
information.
This is supported by ConvoKit in the form of \emph{metadata}.
Each of the four classes in the hierarchy contains a field called \textascii{meta},
which is a lookup table that may be used to store additional information about the \corpus, \conversation, \utterance, or \user under some descriptive name.  
In practice, metadata ranges in complexity from 
speaker
ages to 
sub-utterance level DAMSL speech act tags.

\xhdr{Manipulating conversational data} ConvoKit supports conversational analyses centered on any level of the hierarchy; for instance, one may wish to examine linguistic characteristics of \utterance{s}, characterize a \conversation in terms of the structure of its \utterance{s}, or track a \user's behavior across the \conversation{s} they have taken part in throughout their lifetime. 

Such flexibility in analysis is achieved by abstracting manipulations of conversational data through the \transformer class.
At a high level, a \transformer is an object that takes in a \corpus and returns the same \corpus with some modifications 
applied.
In almost all cases, these modifications will take the form of changed or added metadata. 
For example, the \textascii{PolitenessStrategies} \transformer annotates every \utterance with a feature vector that counts the presence of politeness features from \citet{danescu-niculescu-mizil_computational_2013}, while \textascii{UserConvoDiversity} annotates every \user with a measure of their linguistic diversity across the whole \corpus.

The key to ConvoKit's flexibility is that, while a \transformer can represent any arbitrary manipulation of a \corpus 
and  operate at any level of abstraction,
all \transformer objects share the same syntax---that is, the \transformer class API represents a general language for specifying actions to be taken on a \corpus.
This interface is directly modeled after the scikit-learn class of the same name: a \transformer provides a \fit function and a \transform function.
\fit is used to prepare/train the \transformer with any information it needs beforehand; for example, a \transformer that computes bag-of-words representations of \utterance{s} would first need to build a vocabulary.
\transform then performs the actual modification of the \corpus.

In addition to these standard functions, \transformer{s} also provide a \textascii{summarize()} helper function that offers a high-level tabular or graphical representation of what the \transformer has computed.
For example, \textascii{PolitenessStrategies} offers a \textascii{summarize()} implementation that plots the average occurrence of each politeness feature.
This can be helpful for getting a quick sense of what the \transformer does, for simple exploratory analyses of a \corpus, or for debugging.

A single \transformer on its own might not make significant changes, but because \transformer{s} return the modified \corpus, multiple \transformer{s} can be chained together, each one taking advantage of the previous one's output to produce increasingly complex results
(see Figure \ref{fig:example_code} for an example).

\section{Transformers}
\label{sec:transformers}

\begin{figure}[ht!]
\begin{flushright}
\begin{minipage}{0.93\linewidth}
\begin{lstlisting}[alsoletter={()[]}]
corp = Corpus(filename=download(
  'movie-corpus'))

# Preprocessing step
tc = TextCleaner()
tc.transform(corp)

# Constructing new metadata
for c in corp.iter_conversations():
  genders = [s.meta['gender'] for s in c.iter_speakers()]
  convo.meta['mixed'] = 'M' in genders and 'F' in genders

# Analysis step
fw = FightingWords()
fw.fit(corp, 
  class1_func=lambda utt: utt.get_conversation().meta['mixed'], 
  class2_func=lambda utt: not utt.get_conversation().meta['mixed'])
fw.summarize(corp)
\end{lstlisting}
\end{minipage}
\end{flushright}
\caption{Basic example code demonstrating how combining different \transformer{s}, and leveraging the \corpus hierarchy's rich navigation features and metadata functionality, can be used to study conversational data---in this example, comparing the language used in mixed-gender and single-gender movie dialogs.}
\label{fig:example_code}
\end{figure}

In this section, we introduce some of the built-in \transformer{s} that are available for general use.
Broadly speaking, we can group the functionality of \transformer{s} into three categories: preprocessing, feature extraction, and analysis.

\textbf{Preprocessing} refers to the preliminary processing of the \corpus objects prior to some substantive analysis. For example, at the \utterance-level, preprocessing steps can include 
converting dirty web text into a cleaned ASCII representation (implemented in \textascii{TextCleaner}) or running a dependency parse (implemented in \textascii{TextParser}).
At the \conversation-level, preprocessing steps might include merging consecutive utterances by the same speaker, 
	while  at the \user-level, they might include merging contributions from speakers with multiple user accounts.

\textbf{Feature extraction} 
refers to transformation of conversational data, such as utterance text or conversational structure, into (numerical) features 
for 
further analysis and applications.
An example of an \utterance-level feature extractor is the previously described \textascii{PolitenessStrategies}, while an example of a \conversation-level feature extractor is \textascii{HyperConvo}, which
constructs a hypergraph representation of the \conversation and extracts features such as (generalized) reciprocity, indegree and outdegree distributions, etc.

\textbf{Analysis} is the process of combining  \utterance, \conversation and \user features and metadata into a statistical or machine learning model
to achieve a higher-level understanding of the \corpus. 
For example, \textascii{FightingWords} implements \citet{monroe_fightin_2008}'s method for prinicpled comparison of language used by two subsets of a \corpus; \textascii{Classifier} acts as a wrapper around any scikit-learn machine learning model and can be used to classify \utterance{s}, \conversation{s}, or \user{s} based on the output of feature extraction \transformer{s}; and \textascii{Forecaster} implements \citet{chang_trouble_2019}'s method for modeling the future trajectory of a \conversation.

Figure \ref{fig:example_code} illustrates how \transformer{s} belonging to each category can be combined in sequence to perform a practical conversational task: comparing the language used in movie dialogs containing characters of different genders to that used in dialogs containing only one gender.\footnote{This example, together with its output and other examples, can be found at \url{https://convokit.cornell.edu/documentation/examples.html}.}

\section{Datasets}
\label{sec:data}

ConvoKit ships with a diverse collection of datasets already formatted as \corpus objects and ready for use `out-of-the-box'.
These datasets 
cover the wide range of settings conversational data can come from,
including face-to-face institutional interactions (e.g., supreme court transcripts), 
collaborative online conversations (e.g., Wikipedia talk pages),
threaded social media discussions (e.g., a full dump of Reddit), and even fictional exchanges (e.g., movie dialogs).\footnote{A complete list of datasets can be found at \url{https://convokit.cornell.edu/documentation/datasets.html}.}

The diversity of these datasets further demonstrates the expressiveness of our choice of conversation representation.
We also provide guidelines and code for transforming other datasets into ConvoKit format, allowing ConvoKit's reach to extend beyond what data is already offered.

\section{Conclusions and Future Work}
\label{sec:discussion}

In this paper, we presented ConvoKit, a toolkit that aims to make analysis of conversations accessible to a broad audience.
It achieves this by providing intuitive and user friendly abstractions for both representation and manipulation of conversational data, thus promoting reproducibility and adoption.

ConvoKit is actively being developed. While it is currently heavily centered around text analysis (with other modalities being only indirectly supported as metadata), providing first-class support for spoken dialogs is considered as an important line for future extension. In addition, we aim to continue to incorporate new datasets, analysis methods,  and integrate with other parts of the NLP software ecosystem that could benefit from ConvoKit's abstractions, including dialog generation toolkits such as ParlAI \cite{miller_parlai_2018}.

ConvoKit is an open-source project and we welcome contributions of any kind, ranging from bugfixes and documentation, to augmenting existing corpora with additional useful metadata, to entirely new datasets and analysis methods.\footnote{See contribution guidelines on the ConvoKit webpage.}

\section*{Acknowledgments}

We thank the anonymous reviewers for their thoughtful
comments and are grateful to all ConvoKit contributors. This work was supported by an NSF CAREER award IIS-1750615. Zhang was supported in part by a  Microsoft PhD Fellowship.

\bibliographystyle{acl_natbib}
\bibliography{convokit-sigdial}
\end{document}